\documentclass[conference]{IEEEtran}
\usepackage{blindtext, graphicx}
\usepackage{hyperref}
\hypersetup{
    colorlinks=true,
    linkcolor=blue,
    filecolor=magenta,      
    urlcolor=cyan,
}
 
\ifCLASSINFOpdf
\else
\fi

\begin{document}
%
\title{TernausNet: U-Net with VGG11 Encoder Pre-Trained on ImageNet for Image Segmentation}

\author{
\IEEEauthorblockN{Vladimir Iglovikov}
\IEEEauthorblockA{Lyft Inc.\\
San Francisco, CA 94107, USA\\
Email: iglovikov@gmail.com}
\and
\IEEEauthorblockN{Alexey Shvets}
\IEEEauthorblockA{Massachusetts Institute of Technology\\
Cambridge, MA 02142, USA\\
Email: shvets@mit.edu}
}

\maketitle

\begin{abstract}
Pixel-wise image segmentation is demanding task in computer vision. Classical U-Net architectures composed of encoders and decoders are very popular for segmentation of medical images, satellite images etc. Typically, neural network initialized with weights from a network pre-trained on a large data set like ImageNet shows better performance than those trained from scratch on a small dataset. In some practical applications, particularly in medicine and traffic safety, the accuracy of the models is of utmost importance. In this paper, we demonstrate how the U-Net type architecture can be improved by the use of the pre-trained encoder. Our code and corresponding pre-trained weights are publicly available at https://github.com/ternaus/TernausNet. We compare three weight initialization schemes: LeCun uniform, the encoder with weights from VGG11 and full network trained on the Carvana dataset. This network architecture was a part of the winning solution (1st out of 735) in the Kaggle: Carvana Image Masking Challenge.
\end{abstract}


\IEEEpeerreviewmaketitle

\section{Introduction}
Recent progress in computer hardware with the democratization to perform intensive calculations has enabled researchers to work with models, that have millions of free parameters. Convolutional neural networks (CNN) have already demonstrated their success in image classification, object detection, scene understanding etc. For almost any computer vision problems, CNN-based approaches outperform other techniques and in many cases even human experts in the corresponding field. Now almost all computer vision application try to involve deep learning techniques to improve traditional approaches. They influence our everyday lives and the potential uses of these technologies look truly impressive.

Reliable image segmentation is one of the important tasks in computer vision. This problem is especially important for medical imaging that can potentially improve our diagnostic abilities and in scene understanding to make safe self-driving vehicles. Dense image segmentation essentially involves dividing images into meaningful regions, which can be viewed as a pixel level classification task. The most straightforward (and slow) approach to such problem is manual segmentation of the images. However, this is a time-consuming process that is prone to mistakes and inconsistencies that are unavoidable when human data curators are involved. Automating the treatment provides a systematic way of segmenting an image on the fly as soon as the image is acquired. This process requires providing necessary accuracy to be useful in the production environment.

In the last years, different methods have been proposed to tackle the problem of creating CNN's that can produce a segmentation map for an entire input image in a single forward pass. One of the most successful state-of-the-art deep learning method is based on the Fully Convolutional Networks (FCN) \cite{fcn_2015}. The main idea of this approach is to use CNN as a powerful feature extractor by replacing the fully connected layers by convolution one to output spatial feature maps instead of classification scores. Those maps are further upsampled to produce dense pixel-wise output. This method allows training CNN in the end to end manner for segmentation with input images of arbitrary sizes. Moreover, this approach achieved an improvement in segmentation accuracy over common methods on standard datasets like PASCAL VOC \cite{pascal_voc_2015}. This method has been further improved and now known as U-Net neural network \cite{ronneberger_2015}. The U-Net architecture uses skip connections to combine low-level feature maps with higher-level ones, which enables precise pixel-level localization. A large number of feature channels in upsampling part allows propagating context information to higher resolution layers. This type of network architecture proven themselves in binary image segmentation competitions such as satellite image analysis \cite{iglovikov_2017} and medical image analysis \cite{rakhlin_2017, kalinin_2017} and other \cite{carvana_2017}.

In this paper, we show how the performance of U-Net can be easily improved by using pre-trained weights. As an example, we show the application of such approach to Aerial Image Labeling Dataset \cite{aerialimagelabeling_2017}, that contains aerospace images of several cities with high resolution. Each pixel of the images is labeled as belonging to either "building" or "not-building" classes. Another example of the successful application of such an architecture and initialization scheme is Kaggle Carvana image segmentation competition \cite{carvana_2017}, where one of the authors used it as a part of the winning (1st out 735 teams) solution.

\begin{figure*}
\includegraphics[width=\textwidth,height=12cm]{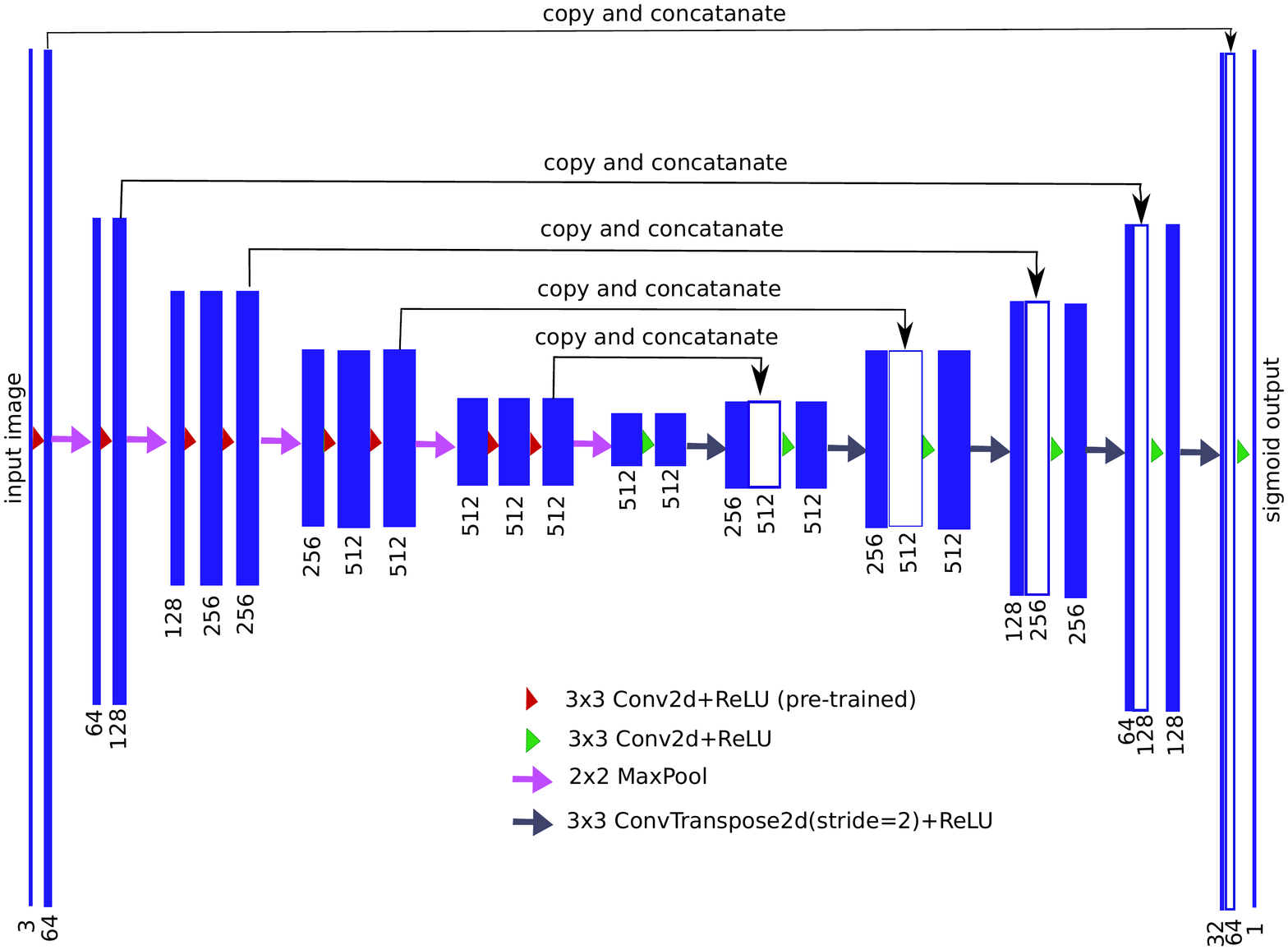}
\caption{Encoder-decoder neural network architecture also known as U-Net where VGG11 neural network without fully connected layers as its encoder. Each blue rectangular block represents a multi-channel features map passing through a series of transformations. The height of the rod shows a relative map size (in pixels), while their widths are proportional to the number of channels (the number is explicitly subscribed to the corresponding rod). The number of channels increases stage by stage on the left part while decrease stage by stage on the right decoding part. The arrows on top show transfer of information from each encoding layer and concatenating it to a corresponding decoding layer.}
\label{fig::unetvgg11}
\end{figure*}

\section{Network Architecture}
In general, a U-Net architecture consists of a contracting path to capture context and of a symmetrically expanding path that enables precise localization (see for example Fig. \ref{fig::unetvgg11}). The contracting path follows the typical architecture of a convolutional network with alternating convolution and pooling operations and progressively downsamples feature maps, increasing the number of feature maps per layer at the same time. Every step in the expansive path consists of an upsampling of the feature map followed by a convolution. Hence, the expansive branch increases the resolution of the output. In order to localize, upsampled features, the expansive path combines them with high-resolution features from the contracting path via skip-connections \cite{ronneberger_2015}. The output of the model is a pixel-by-pixel mask that shows the class of each pixel. This architecture proved itself very useful for segmentation problems with limited amounts of data, e.g. see \cite{iglovikov_2017}.

U-Net is capable of learning from a relatively small training set. In most cases, data sets for image segmentation consist of at most thousands of images, since manual preparation of the masks is a very costly procedure. Typically U-Net is trained from scratch starting with randomly initialized weights. It is well known that training network without over-fitting the data set should be relatively large, millions of images. Networks that are trained on the Imagenet \cite{russakovsky_2014} data set are widely used as a source of the initialization for network weights in other tasks. In this way, the learning procedure can be done for non-pre-trained several layers of the network (sometimes only for the last layer) to take into account features of the date set.

\begin{figure}[ht]
\centering
\includegraphics[width=4cm]{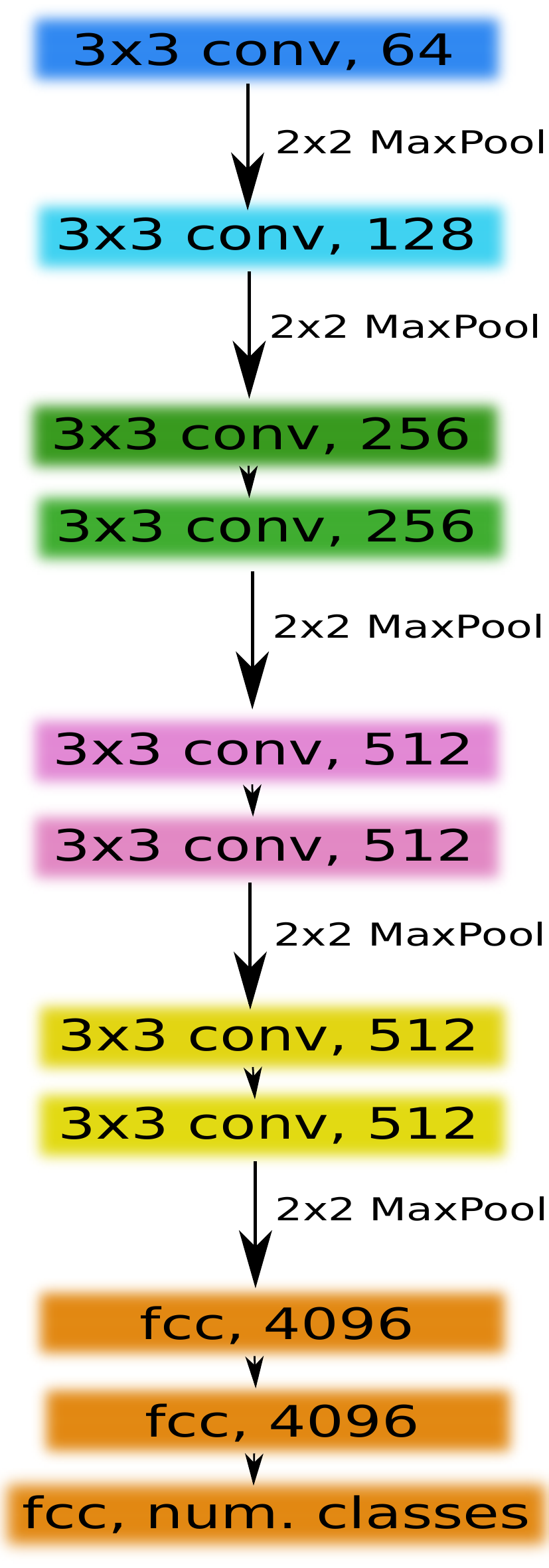}
\caption{VGG11 network architecture. In this picture each convolutional layer is followed by ReLU activation function. The number in each box represents the number of channels in the 
corresponding feature map.}
\label{fig::vgg11}
\end{figure}

As an encoder in our U-Net network, we used relatively simple CNN of the VGG family \cite{vgg_2014} that consists of 11 sequential layers and known as VGG11 see Fig. \ref{fig::vgg11}. VGG11 contains seven convolutional layers, each followed by a ReLU activation function, and five max polling operations, each reducing feature map by $2$. All convolutional layers have $3\times3$ kernels and the number of channels is given in Fig. \ref{fig::vgg11}. The first convolutional layer produces 64 channels and then, as the network deepens, the number of channels doubles after each max pooling operation until it reaches 512. On the following layers, the number of channels does not change.

\begin{figure}[ht]
\centering
\includegraphics[width=9cm]{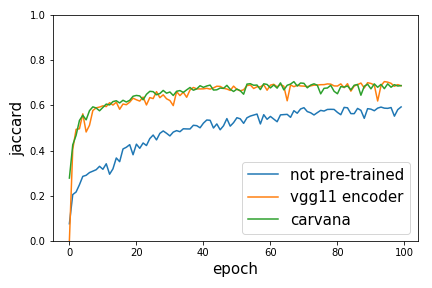}
\caption{Jaccard index as a function of a training epoch for three U-Net models with different weight initialization. The blue line shows a model with randomly initialized weights, orange line shows a model, where the encoder was initialized with VGG11 network pre-trained on ImageNet. Green line shows a model, where the entire network was pre-trained on Carvana data set.}
\label{fig::jaccard}
\end{figure}

To construct an encoder, we remove the fully connected layers and replace them with a single convolutional layer of 512 channels that serves as a bottleneck central part of the network, separating encoder from the decoder. To construct the decoder we use transposed convolutions layers that doubles the size of a feature map while reducing the number of channels by half. And the output of a transposed convolution is then concatenated with an output of the corresponding part of the decoder. The resultant feature map is treated by convolution operation to keep the number of channels the same as in a symmetric encoder term. This upsampling procedure is repeated 5 times to pair up with 5 max poolings, as shown in  Fig. \ref{fig::unetvgg11}. Technically fully connected layers can take an input of any size, but because we have 5 max-pooling layers, each downsampling an image two times, only images with a side divisible by 32 ($2^5$) can be used as an input to the current network implementation.

\begin{figure*}
\includegraphics[width=\textwidth,height=12cm]{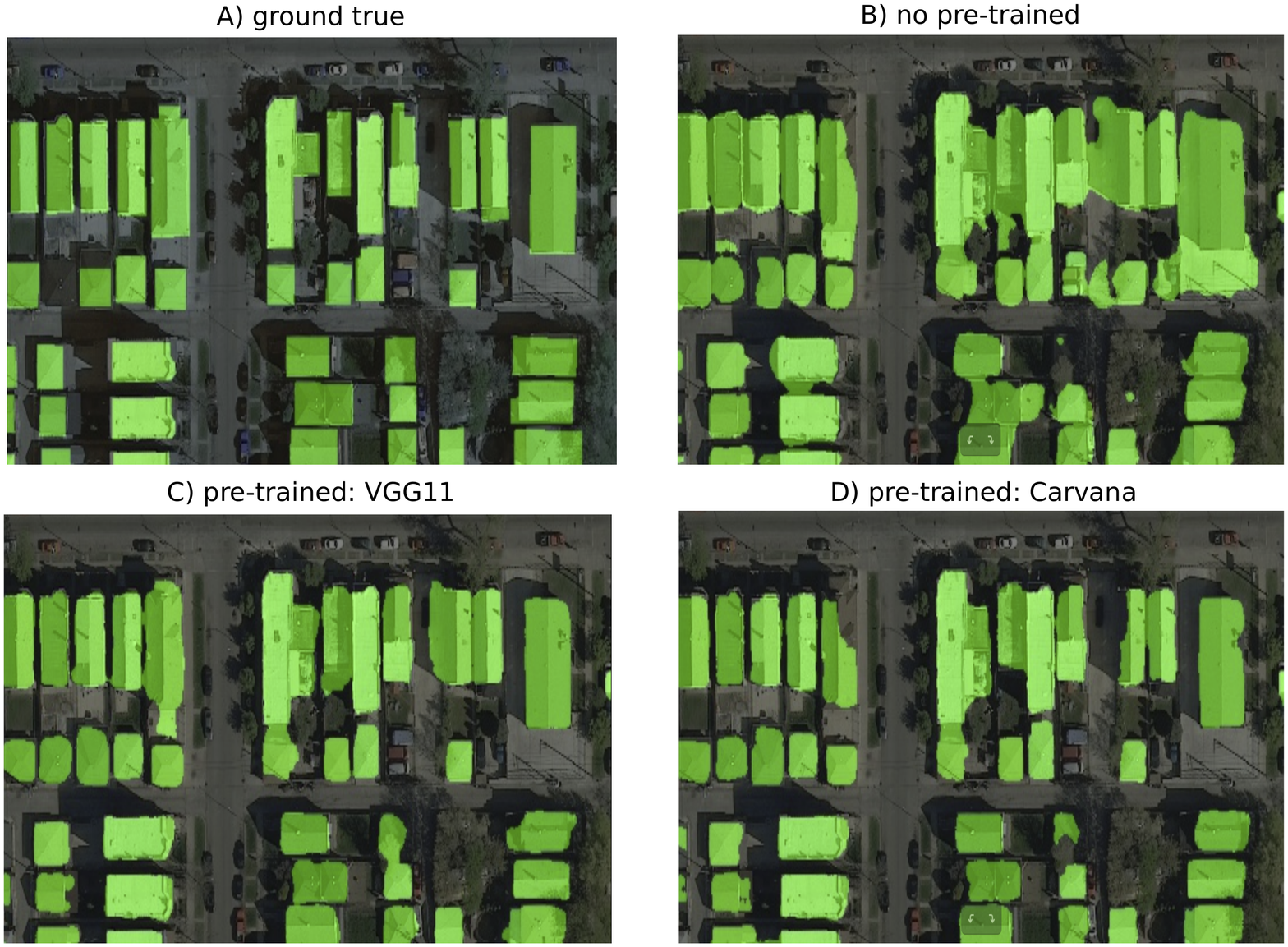}
\caption{Binary masks with green pixels indicate class membership (buildings). Image A) shows an original image with the superimposed ground true mask; Images B) to D) show predictions, initialized with different schemas and trained for 100 epochs. Network in image B) had randomly initialized weights. The model in image C) used randomly initialized decored weights and encoder weights initialized with VGG11, pre-trained on ImageNet. The model in image D) used weights, pre-trained on Carvana data set.}
\label{fig::chicago}
\end{figure*}

\section{Results}
We applied our model to Inria Aerial Image Labeling Dataset \cite{aerialimagelabeling_2017}. This dataset consists of 180 aerial images of urban settlements in Europe and United States, and is labeled as a building and not building classes. Every image in the data set is RGB and has $5000 \times 5000$ pixels resolution where each pixel corresponds to a  $30  \times 30$ cm$^2$ of Earth surface. 
We used 30 images (5 from every 6 cities in the train set) for validation, as suggested in \cite{building_footprints_2017} (valid. IoU $\simeq$ 0.647) and  \cite{inria_label_2017} (best valid. IoU $\simeq$ 0.73) and trained the network on the remaining 150 images for 100 epochs. Random crops of $768 \times 768$ were used for training and central crops $1440 \times 1440$ for validation. Adam with learning rate $0.001$ as an optimization algorithm \cite{adam_2014}. 

We choose Jaccard index (Intersection Over Union) as evaluation metric. It can be interpreted as similarity measure between a finite number of sets. Intersection over union for similarity measure between two sets $A$ and $B$, can be defined as following:
\begin{equation}
\label{jaccard_iou}
    J(A, B) = \frac{|A\cap B|}{|A\cup B|} = \frac{|A\cap B|}{|A|+|B|-|A\cap B|}
\end{equation}
where normalization condition takes place:
$$
    0 \le J(A, B) \le 1
$$
Every image is consists of pixels. To adapt the last expression for discrete objects, we can write it in the following way
\begin{equation}
J=\frac{1}{n}\sum\limits_{i=1}^n\left(\frac{y_i\hat{y}_i}{y_{i}+\hat{y}_i-y_i\hat{y}_i}\right)
\end{equation}
where $y_i$ is a binary value (label) of the corresponding pixel $i$  and $\hat{y}_i$ is predicted probability for the pixel. 

Since, we can consider image segmentation task as a pixel classification problem, we also use the common loss function for binary classification tasks - binary cross entropy that is defined as:

\begin{equation}
H=-\frac{1}{n}\sum\limits_{i=1}^n(y_i\log \hat{y}_i+(1-y_i)\log (1-\hat{y}_i))    
\end{equation}
Join these expressions, we can generalized the loss function, namely, 
\begin{equation}
\label{free_en}
L=H-\log J
\end{equation}

Therefore, minimizing this loss function, we simultaneously maximize probabilities for right pixels to be predicted and maximize the intersection, $J$ between masks and corresponding predictions. For more details, see \cite{iglovikov_2017}.

At the output of a given neural network, we obtain an image where each pixel corresponds to a probability to detect interested area. The size of the output image is coincides with the input image. In order to have only binary pixel values, we choose a threshold 0.3. This number can be found using validation data set and it is pretty universal for our generalized loss function and many different image data sets. For different loss function this number is different and should be found independently. All pixel values below the specified threshold, we set to be zero while all values above the threshold, we set to be 1. Then, multiplying by 255 every pixel in an output image, we can get a black and white predicted mask

In our experiment, we test 3 U-Nets with the same architecture as shown in Fig. \ref{fig::unetvgg11} differing only in the way of weights initialization. For the basic model we use network with weights initialized by LeCun uniform initializer. In this initializer samples draw from a uniform distribution within $[-L, L]$, where $L=\sqrt{1/f_{in}}$ and $f_{in}$ is the number of input units in the weight tensor. This method is implement in pytorch \cite{pytorch} as a default method of weight initialization in convolutional layers. Next, we utilize the same architecture with VGG11 encoder pre-trained on ImageNet while all layers in decoder are initialized by the LeCun uniform initializer. Then, as a final example, we use network with weights pre-trained on Carvana dataset \cite{carvana_2017} (both encoder and decoder). Therefore, after 100 epochs, we obtain the following results for validation subset:

1) LeCun uniform initializer: IoU = 0.593 

2) The Encoder is pre-trained on ImageNet: IoU = 0.686

3) Fully pre-trained U-Net on Carvana: IoU = 0.687

Validation learning curves in Fig. \ref{fig::jaccard} show benefits of our approach. First of all, pre-trained models converge much faster to its steady value in comparison to the non-pre-trained network. Moreover, the steady-state value seems higher for the pre-trained models. Ground truth, as well as three masks, predicted by these three models, are superimposed on an original image in Fig. \ref{fig::chicago}. One can easily notice the difference in the prediction quality after 100 epochs. Moreover, validation learning curves in  Our results for the Inria Aerial Image Labeling Dataset can be easily further improved using different hyper-parameters optimization techniques or standard computer vision methods applying them during pre- and post-processing.

\section{Conclusion}
In this paper, we show how the performance of U-Net can be improved using technique knows as fine-tuning to initialize weights for an encoder of the network. This kind of neural network is widely used for image segmentation tasks and shows state of the art results in many binary image segmentation, competitions. Fine-tuning is already widely used for image classification tasks, but to our knowledge is not with U-Net type family architectures. For the problems of image segmentation, the fine-tuning should be considered even more natural because it is problematic to collect a large volume of training dataset (in particular for medical images) and qualitatively label it. Furthermore, pre-trained networks substantially reduce training time that also helps to prevent over-fitting. Our approach can be further improved considering more advanced pre-trained encoders such as VGG16 \cite{vgg_2014} or any pre-trained network from ResNet family \cite{resnet_2015}. With this improved encoders the decoders can be kept as simple as we use. Our code is available as an open source project under MIT license and can be found at \url{https://github.com/ternaus/TernausNet}.

\section*{Acknowledgment}
The authors would like to thank Open Data Science community \cite{ods} for many valuable discussions and educational help
in the growing field of machine/deep learning. The authors also express their sincere gratitude to Alexander Buslaev who originally suggested to use a pre-trained VGG network as an encoder in a U-Net network.

\ifCLASSOPTIONcaptionsoff
  \newpage
\fi

\begin{IEEEbiography}[{\includegraphics[width=1in,height=1.25in,clip,keepaspectratio]{picture}}]{John Doe}
\blindtext
\end{IEEEbiography}

\end{document}